\documentclass[10pt,twocolumn,letterpaper]{article}

\usepackage{cvpr}
\usepackage{times}
\usepackage{epsfig}
\usepackage{graphicx}
\usepackage{amsmath}
\usepackage{amssymb}

\usepackage{times}
\usepackage{epsfig}
\usepackage{graphicx}
\usepackage{amsmath, amsthm, amssymb, bm}
\usepackage{booktabs}
\usepackage{multirow}
\usepackage{wrapfig}
\usepackage{graphicx}
\usepackage{multirow}
\usepackage{booktabs}
\usepackage{amsmath, amssymb, bm}
\usepackage[pagebackref=true,breaklinks=true,letterpaper=true,colorlinks,bookmarks=false]{hyperref}

\usepackage{colortbl}

\usepackage{makecell, cellspace, caption}
\usepackage[table, svgnames, dvipsnames]{xcolor}
\usepackage[pagebackref=true,breaklinks=true,letterpaper=true,colorlinks,bookmarks=false]{hyperref}

\newcommand{\XX}{\mathbf{X}}
\newcommand{\YY}{\mathbf{Y}}

\newcommand{\yy}{\mathbf{y}}
\newcommand{\pp}{\mathbf{p}}
\newcommand{\qq}{\mathbf{q}}
\newcommand{\PP}{\mathbf{P}}

\newcommand{\real}{\mathbb{R}}

\newcommand{\mypar}[1]{\vspace{.5em}\noindent\textbf{#1}~}

\usepackage[breaklinks=true,bookmarks=false]{hyperref}

\cvprfinalcopy 

\def\cvprPaperID{****} 

\begin{document}

\title{Teach me to segment with mixed supervision:\\ 
Confident students become masters}

\author{Jose Dolz\\
ETS Montreal\\
{\tt\small jose.dolz@etsmtl.ca}
\and
Christian Desrosiers\\
ETS Montreal\\
{\tt\small christian.desrosiers@etsmtl.ca}

\and
Ismail Ben Ayed\\
ETS Montreal\\
{\tt\small ismail.benayed@etsmtl.ca}

}

\maketitle

\begin{abstract}
   Deep neural networks have achieved promising results in a breadth of medical image segmentation tasks. Nevertheless, they require large training datasets with pixel-wise segmentations, which are expensive to obtain in practice. Mixed supervision could mitigate this difficulty, with a small fraction of the data containing complete pixel-wise annotations, while the rest being less supervised, e.g., only a handful of pixels are labeled. In this work, we propose a dual-branch architecture, where the upper branch (teacher) receives strong annotations, while the bottom one (student) is driven by limited supervision and guided by the upper branch. In conjunction with a standard cross-entropy over the labeled pixels, our novel formulation integrates two important terms: (i) a Shannon entropy loss defined over the less-supervised images, which encourages confident student predictions at the bottom branch; and (ii) a Kullback-Leibler  (KL)  divergence, which transfers the knowledge from the predictions generated by the strongly supervised branch to the less-supervised branch, and guides the entropy (student-confidence) term to  avoid trivial solutions. Very  interestingly, we show that the synergy between the entropy and KL divergence yields substantial improvements in performances. Furthermore, we discuss an interesting link between Shannon-entropy minimization  and standard pseudo-mask generation, and argue that the former should be preferred over the latter for leveraging information from unlabeled  pixels. Through a series of quantitative and qualitative experiments, we show the effectiveness of the proposed formulation in segmenting the left-ventricle endocardium in MRI images. We demonstrate that our method significantly outperforms other strategies to tackle semantic segmentation within a mixed-supervision framework. More interestingly, and in line with recent observations in classification, we show that the branch trained with reduced supervision and guided by the top branch largely outperforms the latter.
\end{abstract}

\section{Introduction}

The advent of deep learning has led to the emergence of high-performance algorithms, which have achieved a remarkable success in a wide span of medical image segmentation tasks \cite{litjens2017survey,dolz20183d,ronneberger2015u}. 
One key factor for these advances is the access to large training datasets with high-quality, fully-labeled segmentations. Nevertheless, obtaining such annotations is a cumbersome process prone to observer variability and which, in the case of medical images, requires additional expertise. To alleviate the need for large labeled datasets, weakly supervised learning has recently emerged as an appealing alternative. In this scenario, one has access to a large amount of weakly labeled data, which can come in the form of bounding boxes \cite{kervadec2020bounding,rajchl2016deepcut}, scribbles \cite{lin2016scribblesup,marin2019beyond,tang2018regularized}, image tags \cite{lee2019ficklenet,belharbi2020deep} or anatomical priors \cite{kervadec2019constrained,peng2020discretely,bateson2019constrained}. However, even though numerous attempts have been done to train segmentation models from weak supervision, most of them still fall behind their supervised counterparts, limiting their applicability in real-world settings. 

Another promising learning scenario is mixed supervision, where only a small fraction of data is densely annotated, and a larger dataset contains less-supervised images. In this setting, which enables to keep the annotation budget under control, strongly-labeled data --where all pixels are annotated-- can be combined with images presenting weaker forms of supervision. 
Prior literature \cite{lee2019ficklenet,rajchl2016deepcut} has focused mainly on leveraging weak annotations to generate accurate initial pixel-wise annotations, or \textit{pseudo-masks}, which are then combined with strong supervisions to augment the training dataset. The resulting dataset is employed to train a segmentation network, mimicking fully supervised training. Nevertheless, we argue that treating both equally in a single branch may result in limited improvements, as the less-supervised data is underused. Other approaches resort to multi-task learning \cite{mlynarski2019deep,shah2018ms,wang2019mixed}, where the mainstream task (i.e., segmentation) is assisted by auxiliary objectives that are typically integrated in the form of localization or classification losses.
While multi-task learning might enhance the common representation for both tasks in the feature space, this strategy has some drawbacks. First,  the learning of relevant features is driven by commonalities between the multiple tasks, which may generate suboptimal representations for the mainstream task. Secondly, the specific task-objectives do not allow the direct interaction between the multi-stream outputs. This impedes, for example, explicitly enforcing consistency on the predictions between multiple branches which, as we show in our experiments, significantly improves the results.


Motivated by these observations, we propose a novel formulation for learning with mixed supervision in medical image segmentation. Particularly, our dual-branch network imposes separate processing of the strong and weak annotations, which prevents direct interference of different supervision cues. 
As the uncertainty of the predictions at the unlabeled pixels remains high, we enhance our loss with the Shannon entropy, which enforces high-confidence predictions. Furthermore, in contrast to prior works \cite{mlynarski2019deep,shah2018ms,wang2019mixed}, which have overlooked the co-operation between multiple branches' predictions by considering independent multi-task objectives, we introduce a Kullback-Leibler (KL) divergence term. The benefits of the latter are two-fold. First, it transfers the knowledge from the the predictions generated by the strongly supervised branch (teacher) to the less-supervised branch (student). Second, it guides the entropy (student-confidence) term to avoid trivial solutions.
Very interestingly, we show that the synergy between the entropy and KL term yields substantial improvements in performances. Furthermore, we discuss an interesting link between Shannon-entropy minimization  
and pseudo-mask generation, and argue that the former should be preferred over the latter for leveraging information from unlabeled pixels. We report comprehensive experiments and comparisons with other strategies for learning with mixed supervision, which show the effectiveness of our novel formulation. An interesting finding is that the branch receiving weaker supervision considerably outperforms the strongly supervised branch. This phenomenon, where \textbf{the student surpasses the teacher's performance}, is in line with recent observations in the context of image classification.

\section{Related work}



\mypar{Mixed-supervised segmentation} An appealing alternative to training CNNs with large labeled datasets is to combine a reduced number of fully-labeled images with a larger set of images with reduced annotations. These annotations can come in the form of bounding boxes, scribbles or image tags, for example\footnote{Note that this type of supervision differs from semi-supervised methods, which leverage a small set of labeled images and a much larger set of unlabeled images.}. A large body of the literature in this learning paradigm addresses the problem from a multi-task objective perspective \cite{hong2015decoupled,bhalgat2018annotation,shah2018ms,mlynarski2019deep,wang2019mixed}, which might hinder their capabilities to fully leverage joint information for the mainstream objective. 
Furthermore, these methods typically require carefully-designed task-specific architectures, which also integrate task-dependent auxiliary losses, limiting the applicability to a wider range of annotations. For example, the architecture designed in \cite{shah2018ms} requires, among others, landmark annotations, which might be difficult to obtain in many applications. More recently, Luo et \textit{al.} promoted the use of a dual-branch architecture to deal separately with strongly and weakly labeled data \cite{luo2020semi}. Particularly, while the strongly supervised branch is governed by available fully annotated masks, the weakly supervised branch receives supervision from a proxy ground-truth generator, which requires some extra information, such as class labels. While we advocate the use of independent branches to process naturally different kinds of supervision, we believe that this alone is insufficient, and may lead to suboptimal results. Thus, our work differs from \cite{luo2020semi} in several aspects. First, we make a better use of the labeled images by enforcing consistent segmentations between the strongly and weakly supervised branches on these images. 
Furthermore, we enforce confident predictions at the weakly supervised branch by minimizing the Shannon entropy of the softmax predictions.

\mypar{Distilling knowledge in semantic segmentation} Transferring knowledge from one model to another has recently gained attention in segmentation tasks. For example, the teacher-student strategy has been employed in model compression \cite{bar2019robustness}, to distil knowledge from multi-modal to mono-modal segmentation networks \cite{hu2020knowledge}, or in domain adaptation \cite{xu2019self}. 
Semi-supervised segmentation has also benefited from teacher-student architectures \cite{cui2019semi,sedai2019uncertainty,wang2020self}. In these approaches, however, the segmentation loss evaluating the consistency between the teacher and student models is computed on the unannotated data. A common practice, for example, is to add additive Gaussian noise to the unlabeled images, and enforce similar predictions for the original and noised images. This contrasts with our method, which enforces consistency only on the strongly labeled data, thereby requiring less additional images to close the gap with full supervision.

\begin{figure*}[h!]
    \centering
    \includegraphics[width=0.95\textwidth]{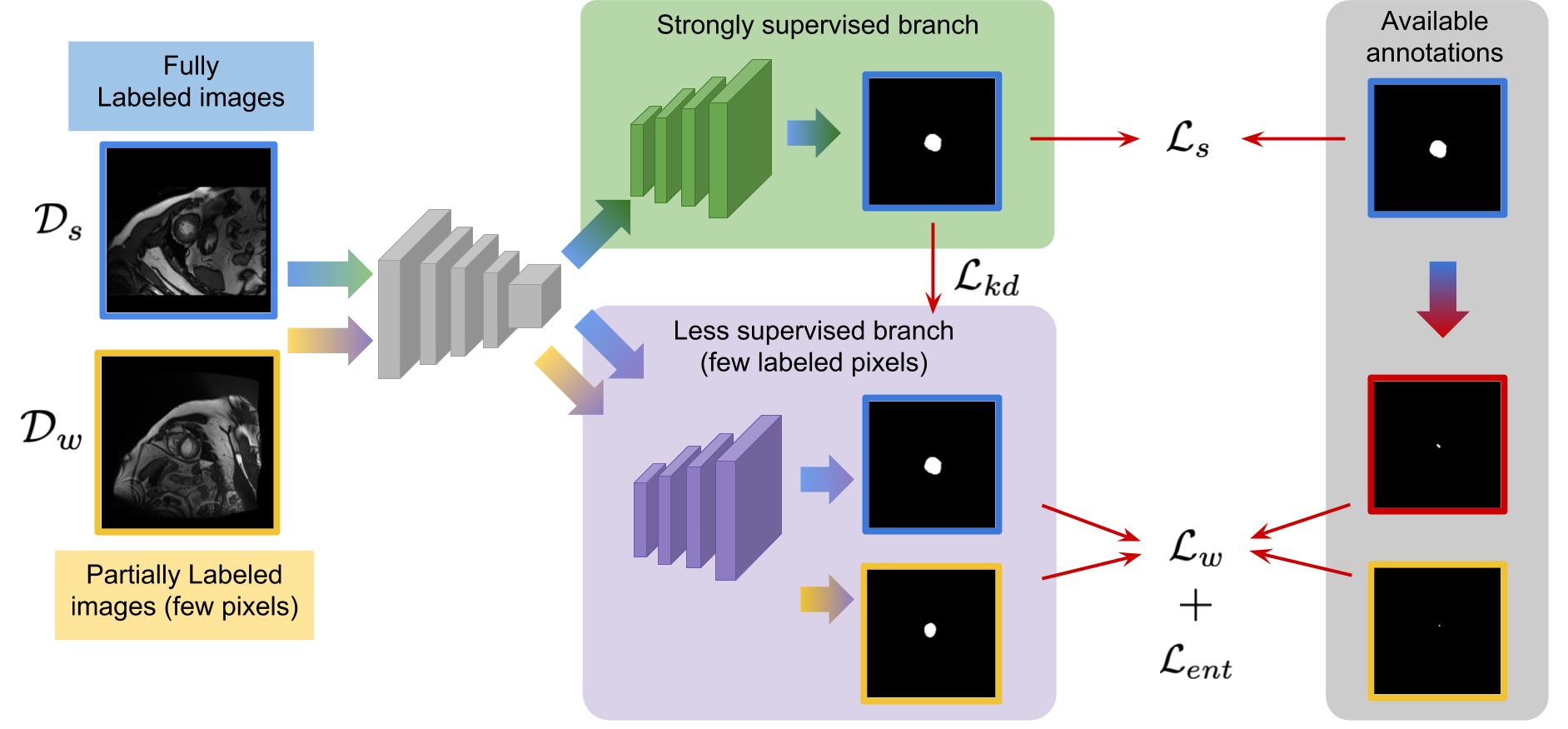}
    \caption[]{Overview of the proposed method. Both fully and partial labeled images are fed to the network. The top branch generates predictions for fully labeled images, whereas the bottom branch generates the outputs for partially labeled images. Furthermore, the bottom branch also generates segmentations for the fully labeled images, which are guided by the KL term between the two branches. } 
    \label{fig:WSL_strategies}
\end{figure*}

\section{Methodology}


Let us denote the set of training images as $\mathcal{D} = \{(\XX_n, \YY_n)\}_n$, where $\XX_i \in \real^{\Omega_i}$ represents the \textit{i}$^{th}$ image and $\YY_i \in \{ 0,1 \}^{\Omega_i \times C}$ its corresponding ground-truth segmentation mask. $\Omega_i$ denotes the spatial image domain and $C$ the number of segmentation classes (or regions). We assume the dataset has two subsets: $\mathcal{D}_s=\{(\XX_1,\YY_1),...,(\XX_m,\YY_m)\}$, which contains complete pixel-level annotations of the associated $C$ categories, and $\mathcal{D}_w=\{(\XX_{m+1},\YY_{m+1}),...,(\XX_n,\YY_n)\}$, whose labels can take the form of semi- or weakly-supervised annotations. Furthermore, for each image $\XX_i$ in $\mathcal{D} = \mathcal{D}_s \cup \mathcal{D}_w$, $\PP_i \in \{ 0,1 \}^{\Omega_i \times C}$ denotes the softmax probability outputs of the network, i.e., the matrix containing a simplex column vector $\pp_i^l = \left ( p_i^{l,1}, \dots, p_i^{l,C} \right )^{T} \in [0, 1]^C$ for each pixel $l \in \Omega_i$. Note that we omit the parameters of the network here to simplify notation.   




\subsection{Multi-branch architecture}

The proposed architecture is composed of multiple branches, each dedicated to a specific type of supervision (see Fig. \ref{fig:WSL_strategies}). It can be divided in two components: a shared feature extractor and independent but identical decoding networks (one per type of supervision), which  differ in the type of annotations received. Even though the proposed multi-branch architecture has similarities with the recent work in \cite{luo2020semi}, there exist significant differences, particularly in the loss functions, which leads to different optimization scenarios. 

\subsection{Supervised learning}

The top-branch is trained under the fully-supervised paradigm, where a set of training images containing pixel-level annotations for all the pixels is given, i.e., $\mathcal{D}_s$. 
The problem amounts to minimizing with respect to the network parameters a standard full-supervision loss, which typically takes the form of a cross-entropy: 
\begin{equation}
\label{eq:CE}
{\cal L}_s = - \sum_{i=1}^m \sum_{l \in \Omega_i} (\yy_i^l)^T  \log \left (\pp_i^l \right )_{\mbox{\tiny top}} 
\end{equation}
where row vector $\yy_i^l = \left ( y_i^{l,1}, \dots, y_i^{l,C} \right ) \in \{0, 1\}^C$ describes the ground-truth annotation for pixel $l \in \Omega_i$. Here, notation $(.)_{\mbox{\tiny top}}$ refers to the softmax outputs of the top branch of the the network.



\subsection{Not so-supervised branch}

We consider the scenario where only the labels for a handful of pixels are known, i.e., scribbles or points. Particularly, we use the dataset $\mathcal{D}_w$ whose pixel-level labels are partially provided. 
Furthermore, for each image on the labeled training set, $\mathcal{D}_s$, we generate partially supervised labels (more details in the experiments' section), which are added to augment the dataset $\mathcal{D}_w$. Then, for the partially-labeled set of pixels, which we denote $\Omega_i^{\mbox{\tiny partial}}$ for each image $i \in \{1, \dots n\}$, we can resort to the following partial-supervision loss, which takes the form of a cross-entropy on the fraction of labeled pixels:
\begin{equation}
\label{eq:CE-bottom}
{\cal L}_w = - \sum_{i=1}^n \sum_{l \in \Omega_i^{\mbox{\tiny partial}}} (\yy_i^l)^T  \log \left (\pp_i^l \right )_{\mbox{\tiny bottom}} 
\end{equation}
where notation $(.)_{\mbox{\tiny bottom}}$ refers to the softmax outputs of the bottom branch of the the network.



\subsection{Distilling strong knowledge}

In addition to the specific supervision available at each branch, we transfer the knowledge from the teacher (top branch) to the student (bottom branch). This is done by forcing the softmax distributions from the bottom branch to mimic the probability predictions generated by the top branch for the fully labeled images in $\mathcal{D}_s$. 
This knowledge-distillation regularizer takes the form of a Kullback-Leibler divergence ($\mathcal{D}_{KL}$) between both distributions:
\begin{equation}
\mathcal{L}_{kd} = \sum_{i=1}^m \sum_{l \in \Omega_i}  \mathcal{D}_{KL}\left ( \left (\pp_i^l \right )_{\mbox{\tiny top}} \| \left (\pp_i^l \right )_{\mbox{\tiny bottom}} \right )
\label{eq:kl}
\end{equation}
where $\mathcal{D}_{KL}(\pp\|\qq)= \pp^T \log \frac{\pp}{\qq}$, with $T$ denoting the transpose operator.

\subsection{Shannon-Entropy minimization}  

Finally, we encourage high confidence in the student softmax predictions for the partially labeled images by minimizing the Shannon entropy of the predictions on the bottom branch:
\begin{equation}
\label{eq:ent-target}
\mathcal{L}_{ent} = \sum_{i=m+1}^n \sum_{l \in \Omega_i}  \mathcal{H} \left ( \pp_i^l \right )
\end{equation}
where $\mathcal{H}\left(\pp \right) = - \pp^T \log \pp$ denotes the Shannon entropy of distribution $\pp$. 

Entropy minimization is widely used in semi-supervised learning (SSL) and transductive classification \cite{grandvalet2005semi,berthelot2019mixmatch,dhillon2019baseline,boudiaf2020information,jabi2019deep} 
to encourage confident predictions at unlabeled data points. Fig. \ref{fig:entropy} plots the entropy in the case of a two-class distribution $(p, 1-p)$, showing how the minimum is reached at the vertices of the simplex, i.e., when $p=0$ or $p=1$. However, surprisingly, in segmentation, entropy is not commonly used, except a few recent works in the different contexts of SSL and domain adaptation \cite{peng2020mutual,bateson2020source,vu2019advent}. As we will see in our experiments, we found that the synergy between the entropy term for confident students, ${\cal L}_{ent}$, and the student-teacher knowledge transfer term, $\mathcal{L}_{kd}$, yield substantial increases in performances. Furthermore, in the following, we discuss an interesting link between {\em pseudo-mask generation}, common in the segmentation literature, and entropy minimization, showing that the former could be viewed as a proxy for minimizing the latter. We further provide insights as to why entropy minimization should be preferred for leveraging information from the set of unlabeled pixels. 

\subsection{Link between entropy and pseudo-mask supervision}  
\begin{figure}[h!]
\includegraphics[width=1\linewidth]{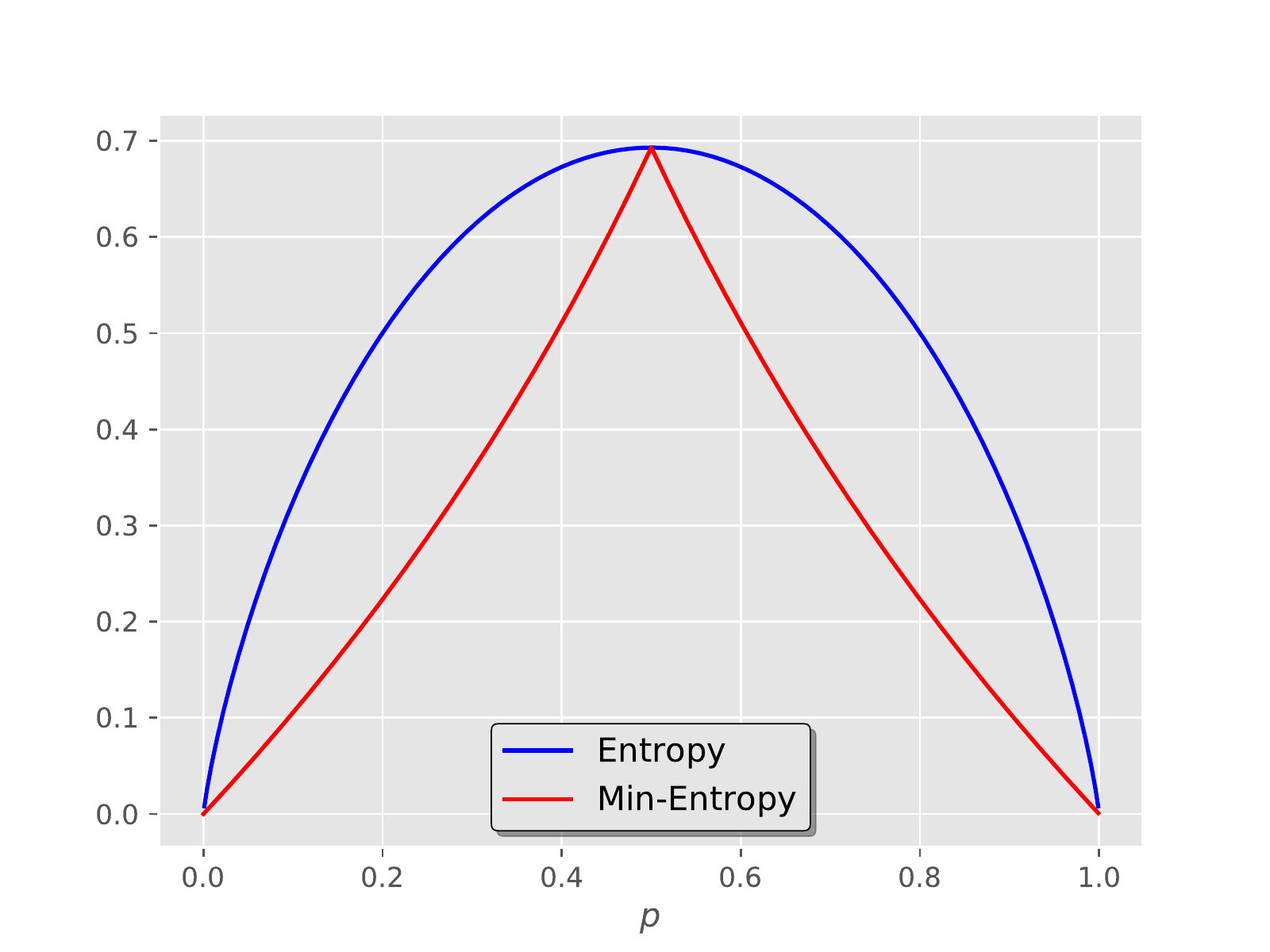}
\caption{Shannon entropy (blue) and min-entropy (red) for a two-class distribution $(p, 1-p)$, with $p \in [0, 1]$.}
\label{fig:entropy}
\end{figure}

In the weakly- and semi-supervised segmentation literature, a very dominant technique to leverage information from unlabeled pixels is to generate pseudo-masks and use these as supervision in a cross-entropy training, in an alternating way \cite{lin2016scribblesup,khoreva2017simple,papandreou2015weakly}. This self-supervision principle is also well known in classification \cite{lee2013pseudo}. Given pixel-wise predictions $\pp_i^l = (p_i^{l,1}, \dots, p_i^{l,C})$, pseudo-masks $q_i^{l,k}$ are generated as follows: $q_i^{l,k} = 1$ if $p_i^{l,k} = \max_{c} p_i^{l,c}$ and $0$ otherwise. By plugging these pseudo-labels in a cross-entropy loss, it is easy to see that this corresponds to minimizing the {\em min-entropy}, ${\mathcal H}_{\mbox{\tiny{min}}}(\pp_i^l) = - \log(\max_{c} p_i^{l,c})$, which is a lower bound on the Shannon entropy; see the red curve in Fig. \ref{fig:entropy}. Fig. \ref{fig:entropy} provides a good insight as to why entropy should be preferred over min-entropy (pseudo-masks) as a training loss for unlabeled data points, and our experiments confirm this.
With entropy, the gradients of un-confident predictions at the middle of the simplex are small and, therefore, dominated by the other terms at the beginning of training. However, with min-entropy, the inaccuracies resulting from un-confident predictions are re-inforced (pushed towards the simplex vertices), yielding early/unrecoverable errors in the predictions, which might mislead training. This is a well-known limitation of self-supervision in the SSL literature \cite{chapelle2009semi}.

\subsection{Joint objective}


Our final loss function takes the following form:
\[\mathcal{L}_{t} = \mathcal{L}_{s} + \lambda_{w} \mathcal{L}_{w} + \lambda_{kd} \mathcal{L}_{kd} + \lambda_{ent} \mathcal{L}_{ent}\]
where $\lambda_{w}$, $\lambda_{kd}$ and $\lambda_{ent}$ balance the importance of each term.

\section{Experimental setting}

\begin{table*}[t!]
\centering
\caption{Results on 
the testing set for the \textit{top} and \textit{bottom} branches (when applicable). Results are averaged over three runs.}

\begin{tabular}{clcc|cc|cc}

\toprule
\multicolumn{1}{l}{}               &           & &    &  \multicolumn{2}{c}{\textit{Top}} & \multicolumn{2}{c}{\textit{Bottom}} \\
\midrule
Setting &     \multicolumn{1}{c}{Model}    & \multicolumn{1}{c}{FS}    & \multicolumn{1}{c}{PS}    & DSC       & HD-95       & DSC     & HD-95    \\
\midrule
\multirow{5}{*}{\textit{Set-3}}  & Lower bound  & $\checkmark$ & -- & 54.66 & 80.05  &  --  &   --  \\
 & \textit{Single}  & $\checkmark$& $\checkmark$&  57.42 & 78.80  & -- &  --  \\
  & \textit{Decoupled} \cite{luo2020semi}     & $\checkmark$ & $\checkmark$ &  56.61 & 74.95   &  5.01 & 120.06  \\
  \rowcolor{Gainsboro!60}
  & \textit{Ours (KL)}  & $\checkmark$& $\checkmark$ & 56.39 & 63.27  &   63.98 &  67.67   \\
  \rowcolor{Gainsboro!60}
  & \textit{Ours (KL+Ent)}   & $\checkmark$& $\checkmark$ &  69.25  & 49.93  & \bf 75.92  & \bf 30.12   \\
\midrule
\multirow{5}{*}{\textit{Set-5}}  & Lower bound  & $\checkmark$ & -- & 69.71&  51.75   &  --  &   --  \\
 & \textit{Single}  & $\checkmark$& $\checkmark$   & 70.73  & 51.34 & -- &  --  \\
  & \textit{Decoupled} \cite{luo2020semi}     & $\checkmark$ & $\checkmark$ & 70.96   & 54.42   &  4.29 &  127.68\\
  \rowcolor{Gainsboro!60}
  & \textit{Ours (KL)}  & $\checkmark$& $\checkmark$ &  67.96 &  44.01  &  72.69 & 40.75   \\
  \rowcolor{Gainsboro!60}
  & \textit{Ours (KL+Ent)} & $\checkmark$& $\checkmark$  &67.10 & 45.28  &  \bf 78.77 & \bf 23.29   \\
\midrule
\multirow{5}{*}{\textit{Set-10}} & Lower bound  & $\checkmark$ & -- & 78.28&  44.16    &   --  &   -- \\ 
& \textit{Single}  & $\checkmark$& $\checkmark$& 78.17  & 42.99   &  -- & --  \\
 & \textit{Decoupled} \cite{luo2020semi}     & $\checkmark$ & $\checkmark$ &  77.53 & 32.23   &  4.58 &  125.36 \\
 \rowcolor{Gainsboro!60}
  & \textit{Ours (KL)}  & $\checkmark$& $\checkmark$ &80.60 &  27.19 &    82.21  &  33.96   \\
  \rowcolor{Gainsboro!60}
  & \textit{Ours (KL+Ent)}  & $\checkmark$& $\checkmark$ & 83.96 &  30.71 &   \bf 88.07  & \bf 4.37  \\
  \midrule
 All images  & Upper bound  & $\checkmark$ & -- & 93.31  &  3.46  &  -- & --  \\
  \bottomrule
  \multicolumn{8}{l}{\scriptsize{FS and PS indicate  full or partial supervised images.}}\\
\end{tabular}

\label{table:acdc-main}
\end{table*}

\mypar{Benchmark dataset}
We focused on the task of left ventricular (LV) endocardium segmentation on cine MRI images. Particularly, we used the training set from the publicly available data of the 2017 ACDC Challenge \cite{bernard2018deep}, which consists of 100 cine magnetic resonance (MR) exams covering several well defined pathologies, 
each exam containing acquisitions only at the diastolic and systolic phases. 
We split this dataset into 80 exams for training, 5 for validation and the remaining 15 for testing. 

\mypar{Generating partially labeled images} The training exams are divided into a small set of fully labeled images, $\mathcal{D}_s$, and a larger set of images with reduced supervision, $\mathcal{D}_w$, where only a handful of pixels are labeled. Concretely, we employ the same partial labels as in \cite{kervadec2019curriculum,kervadec2019constrained}. To evaluate how increasing the amount of both fully and partially labeled affects the performance, we evaluated the proposed models in three settings, referred to as \textit{Set-3}, \textit{Set-5}, and \textit{Set-10}. In these settings, the number of fully labeled images is 3, 5 and 10, respectively, while the number of images with partial labels is $\times$5 times the number of labeled images.

\mypar{Evaluation metrics} For evaluation purposes we employ two well-known metrics in medical image segmentation: the Dice similarity score (DSC) and the modified Hausdorff-Distance (MHD). Particularly, the MHD represents the 95\textit{th} percentile of the symmetric HD between the binary objects in two images.

\mypar{Baseline methods}To demonstrate the efficiency of the proposed model, we compared to several baselines. First, we include full-supervised baselines that will act as lower and upper bounds. The lower bound employs only a small set of fully labeled images (either 3, 5 or 10, depending on the setting), whereas the upper bound considers all the available training images. Then, we consider a single-branch network, referred to as \textit{Single}, which receives both fully and partial labeled images without making distinction between them. To assess the impact of decoupling the branches without further supervision, similar to \cite{luo2020semi}, we modify the baseline network by integrating two independent decoders, while the encoder remains the same. This model, which we refer to as \textit{Decoupled}, is governed by different types of supervision at each branch. Then, our first model, which we refer to as \textit{KL}, integrates the KL divergence term presented in Eq. (\ref{eq:kl}). Last, \textit{KL+Ent} corresponds to the whole proposed model, which couples the two important terms in Eq. (\ref{eq:kl}) and Eq. (\ref{eq:ent-target}) in the formulation.

\mypar{Implementation details} We employed UNet as backbone architecture for the single branch models. Regarding the dual-branch architectures, we modified the decoding path of the standard UNet to accommodate two separate branches. All the networks are trained during 500 epochs by using Adam optimizer, with a batch size equal to 8. We empirically set the values of $\lambda_{w}$, $\lambda_{kd}$ and $\lambda_{ent}$ to 0.1, 50 and 1, respectively. We found that our formulation provided the best results when the input distributions to the KL term in eq. (\ref{eq:kl}) were very smooth, which was achieved by applying softmax over the softmax predictions. All the hyperparameters were fixed by using the independent validation set. Furthermore, we perform 3 runs for each model and report the average values. The code was implemented in PyTorch and all the experiments were performed in a server equipped with a NVIDIA Titan RTX. The code and trained models are publicly available at \href{https://github.com/josedolz/MSL-student-becomes-master}{https://github.com/josedolz/MSL-student-becomes-master}


\subsection{Results}

\paragraph{\textbf{Main results}} Table \ref{table:acdc-main} reports the quantitative evaluation of the proposed method compared to the different baselines. 
First, we observe that across all the settings, simply adding partial annotations to the training set does not considerably improve the segmentation performance. Nevertheless, by integrating the guidance from the upper branch the network is capable of leveraging additional partially-labeled images more efficiently, through the bottom branch. Furthermore, if we couple the KL divergence term with an objective based on minimizing the entropy of the predictions on the partial labeled images, the segmentation performance substantially increases. Particularly, the gain obtained by the complete model is consistent across the several settings, improving the DSC by 6-12\% compared to the \textit{KL} model, and reducing the MHD by nearly 30\%. Compared to the baseline dual-branch model, i.e., \textit{Decoupled}, our approach brings improvements of 10-20\% in terms of DSC and reduces the MHD values by 30-40\%. These results demonstrate the strong capabilities of the proposed model to leverage fully and partially labeled images during training. It is noteworthy to mention that findings on these results, where \textbf{the student excels the teacher}, aligns with recent observations in classification \cite{furlanello2018born,yim2017gift}.

\paragraph{\textbf{Comparison with proposals}}

As mentioned previously, a popular paradigm in weakly and semi-supervised segmentation is to resort to pseudo-masks generated by a trained model, which are used to re-train the network mimicking full supervision. 
To demonstrate that our model leverages more efficiently the available data, we train a network with the proposals generated by the \textit{Lower bound} and \textit{KL} models, whose results are reported in Table \ref{table:acdc-kl}. We can observe that despite typically improving the base model, minimizing the cross-entropy over proposals does not outperform directly minimizing the entropy on the predictions of the partially labeled images. 

\begin{table}[ht!]
\footnotesize
\centering
\caption{Results obtained by training on an augmented dataset composed by fully labeled images and proposals generated from the \textit{Lower bound} and \textit{KL} models (Results obtained by both are reported in Table \ref{table:acdc-main}).}
\begin{tabular}{l|cc|cc|cc}

\toprule

\multicolumn{1}{l}{} & \multicolumn{2}{c}{\textit{\begin{tabular}[c]{@{}c@{}}Proposals\\ (Lower bound)\end{tabular} }} & \multicolumn{2}{c}{\textit{\begin{tabular}[c]{@{}c@{}}Proposals\\ (KL)\end{tabular} }} & \multicolumn{2}{c}{\textit{\begin{tabular}[c]{@{}c@{}}Ours\\ (KL+Ent)\end{tabular} }} \\
\midrule
\multicolumn{1}{l}{Setting}&  DSC       & HD-95  & DSC       & HD-95   & DSC       & HD-95          \\
\midrule
 \textit{Set-3}    & 63.11 & 49.99 & 70.94 &  45.32 & \cellcolor[HTML]{FFFFFF}{\bf 75.92} & \cellcolor[HTML]{FFFFFF}{\bf 30.12}  \\
 \textit{Set-5}    & 73.91 & 45.54& 75.06 &  40.62 & \cellcolor[HTML]{FFFFFF}{\bf 78.77} & \cellcolor[HTML]{FFFFFF}{\bf 23.29}  \\
 \textit{Set-10}    & 81.31 & 29.95 & 82.78  & 24.36  & \cellcolor[HTML]{FFFFFF}{\bf 88.07} & \cellcolor[HTML]{FFFFFF}{\bf 4.37}  \\
  \bottomrule
\end{tabular}
\label{table:acdc-kl}
\end{table}

\paragraph{\textbf{Ablation study on the importance of the KL term}}

The objective of this ablation study is to assess the effect of balancing the importance of the KL term in our formulation. Particularly, the KL term plays a crucial role in the proposed formulation, as it guides the entropy term during training to avoid degenerate solutions. We note that the value of the KL term is typically 2 orders of magnitude smaller than the entropy objective. Therefore, by setting its weight ($\lambda_{kl}$) to 1, we demonstrate empirically its crucial role during training when coupled with the entropy term, as in this setting the latter strongly dominates the training. In this scenario, we observe that the model is negatively impacted, particularly when fully-labeled images are scarce, i.e., \textit{Set-3}, significantly outperforming the lower bound model. This confirms our hypothesis that minimizing the entropy alone results in degenerated solutions. Increasing the weight of the KL term typically alleviates this issue. However, if much importance is given to this objective the performance also degrades. This is likely due to the fact that the bottom branch is strongly encouraged to follow the behaviour of the top branch, and the effect of the entropy  term is diminished.

\begin{table}[t!]
\scriptsize
\centering
\caption{Impact of $\lambda_{kl}$ on the proposed formulation.}\label{tab:results-acdc-ablationK}
\begin{tabular}{l|cc|cc|cc|}
\toprule
         & \multicolumn{2}{c}{\textit{Set-3}} & \multicolumn{2}{c}{\textit{Set-5}} & \multicolumn{2}{c}{\textit{Set-10}} \\
         \midrule
         & DSC        & HD-95       & DSC        & HD-95       & DSC        & HD-95        \\
\midrule
$\lambda_{K}=1$  & 24.89 & 117.52  & 46.30 & 82.89 & 73.88 & 38.23 \\
$\lambda_{K}=10$  & 64.42  & 67.47  & 59.16 & 58.01& 78.66 & 31.79\\
$\lambda_{K}=20$  & 72.30  & 47.52 & 70.47 & 37.84 & 83.47 & 16.97 \\
$\lambda_{K}=50$    & \bf 75.92  & \bf 30.12 & \bf 78.77 & \bf 23.29 & \bf 88.07   & \bf 4.37  \\
$\lambda_{K}=100$   & 60.64  &  71.62 & 66.01 & 43.83& 84.34 & 16.60 \\    
\bottomrule
\end{tabular}

\label{tab:results-acdc-ablationK}
\end{table}

\begin{figure*}[h!]
    \centering
    \includegraphics[width=0.9\textwidth]{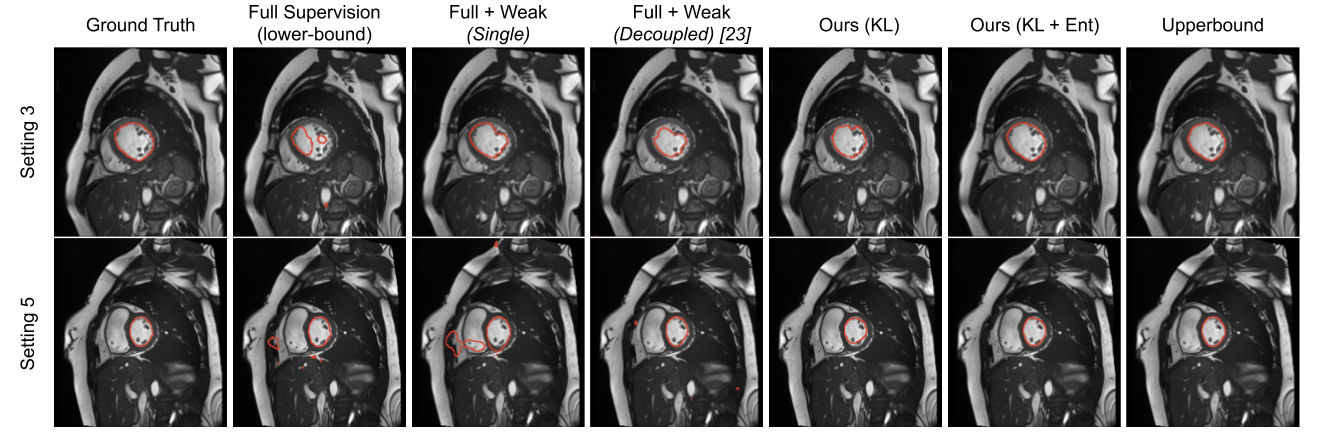}
    \caption[]{Qualitative results for the analyzed models under two different settings.}
    \label{fig:qualitative-main}
\end{figure*}
\begin{figure*}[h!]
    \centering
    \includegraphics[width=0.6\textwidth]{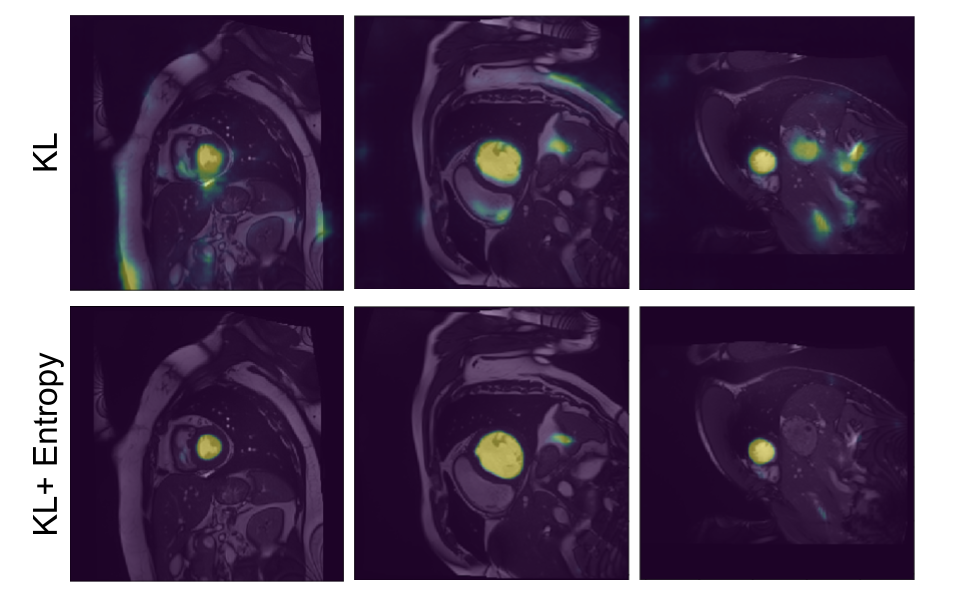}
    \caption[]{\vspace{-1mm}Probability maps obtained by the proposed $KL$ and $KL+Ent$ models.}
    \label{fig:KL-vs-Entropy}
\end{figure*}

\mypar{Qualitative results} In addition to the numerical results presented, we also depict qualitative results in Fig. \ref{fig:qualitative-main} and Fig. \ref{fig:KL-vs-Entropy}. Particularly, Fig. \ref{fig:qualitative-main} depicts the segmentation results for the models evaluated in Table \ref{table:acdc-main}. We can observe that segmentation results obtained by models with a single network typically undersegment the object of interest (\textit{first row}) or generate many false positives (\textit{second row}). Decoupling the decoding branches might reduce the false positive rate, however, it also tends to undersegment the target. Finally, we observe that both of our formulations achieve qualitatively better segmentation results, with the \textit{KL+Ent} model yielding segmentations similar to those generated by the upper bound model. Furthermore, in Fig. \ref{fig:KL-vs-Entropy} we illustrate additional qualitative results of our models. We observe that without the entropy term our model produces less confident predictions, which results in more noisy segmentations.

{\small
\bibliographystyle{ieee_fullname}
\bibliography{egbib}
}

\end{document}